\documentclass{article}

\PassOptionsToPackage{numbers, compress}{natbib}
\usepackage[preprint]{nips_2018}

\usepackage[utf8]{inputenc} % allow utf-8 input
\usepackage[T1]{fontenc}    % use 8-bit T1 fonts
\usepackage{hyperref}       % hyperlinks
\usepackage{url}            % simple URL typesetting
\usepackage{booktabs}       % professional-quality tables
\usepackage{amsfonts}       % blackboard math symbols
\usepackage{nicefrac}       % compact symbols for 1/2, etc.
\usepackage{microtype}      % microtypography
\usepackage{amsmath}
\usepackage{color}
\usepackage{csquotes}

\usepackage{graphicx}
\graphicspath{ {images/} }
\usepackage{scrextend}
\usepackage{booktabs}
\usepackage{float}
\usepackage{listings}

\usepackage{tikz}
\usetikzlibrary{math} 
\usetikzlibrary{shapes,arrows}
\usetikzlibrary{positioning}
\usetikzlibrary{arrows.meta} 

\usepackage{subcaption}

\usepackage{amsthm}

\usepackage{algorithm}
\usepackage{algorithmicx}
\usepackage[noend]{algpseudocode}

\usepackage{relsize}

\usepackage{titlesec}
\titlespacing*{\section}{0pt}{1em}{1em}
\titlespacing*{\subsection}{0pt}{1em}{1em}

\makeatletter
\def\BState{\State\hskip-\ALG@thistlm}
\makeatother

\tikzstyle{block} = [draw,fill=blue!20,minimum size=2em]
\tikzstyle{branch}=[fill,shape=circle,minimum size=3pt,inner sep=0pt]

\definecolor{xuxi}{rgb}{0.59, 0.0, 0.09}
\definecolor{josh}{rgb}{0.0, 0.42, 0.24}
\definecolor{proposed}{rgb}{1.0, 0.49, 0.0}
\definecolor{accepted}{rgb}{0.0, 0.0, 0.0} 

\newcommand{\exact}{\texttt{Exact} }
\newcommand{\algo}{\texttt{Memoryless} }

\title{Memoryless Exact Solutions for Deterministic MDPs with Sparse Rewards}

\author{
  Joshua R. Bertram    \quad    Peng Wei\\
  Iowa State University\\
  Ames, IA 50011 \\
  \texttt{\{bertram1, pwei\}@iastate.edu} \\
}

\begin{document}
% \nipsfinalcopy is no longer used

\maketitle

\begin{abstract}
    We propose an algorithm for deterministic continuous Markov Decision Processes with sparse rewards that computes the optimal policy exactly with no dependency on the size of the state space.  The algorithm has time complexity of $O( |R|^3 \times |A|^2 )$ and memory complexity of $O( |R| \times |A| )$, where $|R|$ is the number of reward sources and $|A|$ is the number of actions.  Furthermore, we describe a companion algorithm that can follow the optimal policy from any initial state without computing the entire value function, instead computing on-demand the value of states as they are needed.  The algorithm to solve the MDP does not depend on the size of the state space for either time or memory complexity, and the ability to follow the optimal policy is linear in time and space with the path length of following the optimal policy from the initial state.  We demonstrate the algorithm operation side by side with value iteration on tractable MDPs.
    %, and demonstrate the new algorithm in operation on exceedingly large state spaces designed to be intractable.
\end{abstract}

\section{Introduction}
	Markov Decision Processes (MDPs) are a framework for decision making with broad applications to financial, robotics, operations research and many other domains.  At their root, MDPs are formulated as the tuple $S, A, R, T $ where $S$ is the state at a given time $t$, $A$ is the action taken by the agent at time $t$ as a result of the decision process, $R$ is the reward received by the agent as a result of taking the action, and $T(s, a, s')$ is a transition function that describes the dynamics of the environment and capture the probability $p( s' | s, a )$ of transitioning to a state $s'$ given the action $a$ taken from state $s$.  An MDP is said to be deterministic if there is no uncertainty or randomness on the transition between $s$ and $s'$ given $a$.  The output of an MDP is termed a policy, $\pi$, which describes an action $a$ that should be taken at every state $s \in S$.  When an MDP is solved completely such that the policy is optimal, it is typically denoted as $\pi^*$.  The optimal policy has the property that it maximizes the expected cumulative reward from any initial starting state.  Alternatively, the MDP solution can also be viewed as a value function that describes the value of being at each state, or also as a $Q$-value function that describes the value of taking a specific action from a given state.  Given one representation, it is possible to recover the other representations.  We use the notation of $V$ for the value function and $V^*$ for the optimal value function.  MDPs which contain states which "terminate", meaning that once the state is reached, no further actions are taken.  In chess, for example, a terminating state would be checkmate.  In other problems there may be no natural terminating state.  Such problems are said to be continuous.  
    
    The reward function $R$ defines the reward that the agent receives for taking action $a$ from state $s$.  Reward functions can be based off only the state, $R(s)$, off the state and action, $R(s, a)$, and occasionally off the resulting next state, $R(s, a, s')$.  
    
    There are many well known methods for solving MDPs exactly including value iteration and policy iteration, which are iterative methods based on the dynamic programming approach proposed by Bellman.  These algorithms use a table-based approach to represent the state-action space exactly and iteratively converge to the optimal policy $\pi^*$ and corresponding value function $V^*$.  These table-based methods have a well known disadvantage that they quickly become intractable.  As the number of states and actions increases in number or dimension, the number of entries in the (multi-dimensional) table increases exponentially.  Many real-world problems quickly exhaust the resources of even high performing computers.
    
    This curse of dimensionality is typically overcome by resorting to various forms of approximation of the optimal value function or optimal policy, some of which also have convergence guarantees or bounds on the error.  Other techniques have focused on managing the size of the state space explosion through factorization or through aggregation and tiling.
    
    Bertram \cite{bertram} proposes an algorithm to solve MDPs named \exact that treats an MDP as a graph and uses the connectivity of the graph and the distance between nodes in the graph to solve an MDP in $O( |R|^2 \times |A|^2 \times |S|)$ time complexity and $O( |S| + |R| \times |A| )$ memory complexity.  The output of the algorithm was the full table-based representation of the value function $V^*$.  This paper was restricted to deterministic continuous MDPs and provided performance improvements when the number of rewards was small compared to the number of states ($ |R| << |S| $) and only supported reward functions based on state, $R(s)$.  While performance was improved for this small class of MDPs, it continued to have a dependence on the size of the state space $|S|$ and therefore ultimately only partially mitigates the curse of dimensionality.  It may be able to support larger state spaces than typical value iteration or policy iteration, but it eventually too will become intractable as the state action space itself grows exponentially -- even linear dependence on the size of the state action space is a significant limitation.  
    
    In this paper, we propose an extension to \exact which we name \algo that \textbf{removes the dependence on the size of the state space} resulting in time complexity of $O( |R|^3 \times |A|^2 )$ and memory complexity of $O( |R| \times |A| )$ for the same restricted class of MDPs.  Rather than outputting the full value function, the algorithm in this paper outputs an ordered list in which rewards should be processed using the same techniques as described in \cite{bertram}.  We propose a companion algorithm that can efficiently follow the optimal policy by calculating the value of neighboring states on-demand.  We show performance against both value iteration and the prior algorithm for tractable state spaces.   %We demonstrate our algorithm performance on a contrived state space that is meant to far exceed the limits of memory to demonstrate the scaling of the algorithm, and we demonstrate the performance of following the optimal policy on this otherwise intractable MDP formulation.
    
\section{Related Work}

Dealing with the 'curse of dimensionality' \cite{bellman2013dynamic} has been an ongoing struggle within the machine learning and optimization communities for many decades, especially within the Markov Decision Process community.  Many attempts have been made to allow MDPs to scale to larger problems.  Factored MDPs \cite{schuurmans2002direct,guestrin2003efficient}  attempt to alleviate the problem of state space explosion by identifying subsets of the MDP that can be broken into smaller problems.  Primarily, though, approximation methods under the general umbrella of Approximate Dynamic Programming have been used as a compromise to obtain reasonable approximations of the underlying true value function in cases where the state-action space (or the transition matrix $T$) is too large to represent with traditional exact methods, which are summarized by the excellent texts \cite{bertsekas1995dynamic,powell2007approximate}.  Notably, linear function approximation methods such as GradientTD methods \cite{gradientTD,tdcandgtd2,precup2001off}, statistical evaluation methods such montecarlo tree search \cite{kocsis2006bandit}, and non-linear function approximation methods such as TDC with non-linear function approximation \cite{bhatnagar2009convergent} and DQN \cite{DQNatari} are good examples of some of the approaches taken using approximation.

\cite{bertram} identifies a new method for a restricted class of MDPs that solves MDPs exactly by calculating the effect of each reward in the state space based off the distance to each reward.  This paper relies heavily on that work and proposes an optimization to that algorithm that removes the dependency on the size of the state space.
    
\section{Methodology}

    See \cite{bertram} for details, but here we briefly recall for the reader that the key insight from this work was to describe an MDP in terms of a graph, to take advantage of known structure of the MDP, and to utilize discoveries on how the expected reward from multiple reward sources interplay and result in the value function.  
    
    The algorithm locates "peaks" in the value function due to the collection of rewards.  It iteratively selects the most valuable peak and calculates an intermediate value function which represents the optimal solution of an MDP with the same environment but a subset of the rewards.  The iterations continue until all rewards have been considered, and results in the optimal value function for the original MDP. 

% Define block styles
\tikzstyle{block} = [rectangle, draw, fill=blue!20, 
    text width=6em, font=\relsize{2}, text centered, rounded corners, minimum height=4em]
\tikzstyle{line} = [draw, -latex', line width=1mm]
\tikzstyle{cloud} = [draw, text centered, ellipse,fill=red!20, node distance=3cm, font=\relsize{2},
    minimum height=2em, text width=6.5em]

\begin{figure}[H]
\centering
\begin{subfigure}{.5\textwidth}
  \centering
\resizebox{\textwidth}{!}{
\begin{tikzpicture}[node distance = 4cm, auto]
    % Place nodes
    \node [cloud] (A) {rewards};
    \node [cloud, right=1cm of A] (B) {peaks};
    \node [block, below=1cm of B] (C) {process peaks};
    \node [cloud, below=1cm of C] (D) {intermediate value function};
    \node [cloud, right=1cm of C] (E) {final value function};
    % Draw edges
    \path [line] (A) -- (B);
    \path [line] (B) -- (C);
    \path [line] (C) -- (D);
    \path [line] (C) -- (E);
    \path [line] (D.west) --  ++(-1,0) |-  (C.west);
\end{tikzpicture}
}
  \caption{\exact algorithm from \cite{bertram}.}
  \label{fig:rewards}
\end{subfigure}%
\begin{subfigure}{.5\textwidth}
  \centering
\resizebox{\textwidth}{!}{
\begin{tikzpicture}[node distance = 4cm, auto]
    % Place nodes
    \node [cloud] (A) {rewards};
    \node [cloud, right=1cm of A] (B) {peaks};
    \node [block, below=1cm of B] (C) {process peaks};
    \node [cloud, below=1cm of C] (D) {ordered list of peaks};
    \node [block, left=1cm of D] (F) {compute value of neighbors};
    \node [cloud, right=1cm of C] (E) {final list of peaks};
    % Draw edges
    \path [line] (A) -- (B);
    \path [line] (B) -- (C);
    \path [line] (C) -- (D);
    \path [line] (C) -- (E);
    \path [line] (D) -- (F);
    \path [line] (F.north) |-  (C.west);
\end{tikzpicture}
}

  \caption{\algo algorithm proposed in this paper.}
  \label{fig:states}
\end{subfigure}
\caption{Changes in algorithm from computing an intermediate value function to computing states on-demand from an ordered list of peaks, obviating the need for the intermediate value function.}
\label{fig:test}
\end{figure}
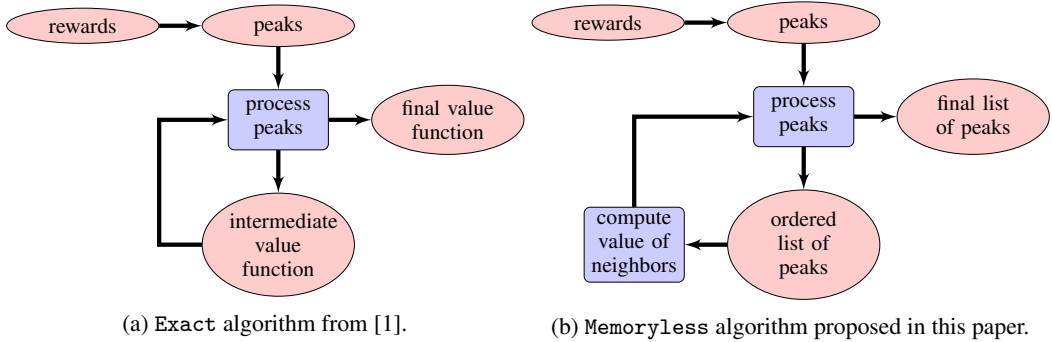

    In the proof for the algorithm in \cite{bertram}, it was shown that the complete value function can be determined from these peaks.  As the algorithm processes each peak, it examines neighboring states, referring to the intermediate value function to look up values of these neighboring states.  Note however that the number of neighboring states that are looked up is typically a very small number (on the order of $O(|R| \times |A|)$); in essence, because the values of a few states are needed, the values of all states are computed each iteration of the algorithm.
    
    Instead, this paper changes the algorithm to compute the neighboring state values as needed from a list of the peaks sorted by order in which they were processed by the algorithm.  Additionally, this paper proposes a mechanism to calculate the value of any state from this ordered list.  During each iteration of the algorithm, this method calculates the value of any required states and results in a final list of how all peaks were processed.

This change to the algorithm severs its dependence on the size of the state space $|S|$, effectively trading between additional computation time and memory storage.  The intermediate value function can be viewed as a lookup table that improves computational efficiency; the new method essentially sacrifices this lookup table method for a slower computation based method that requires a pass through the list of peaks, a $O(|R|)$ operation.  However, when the number of rewards is small, this tradeoff can be acceptable, especially considering that the algorithm is no longer dependent on the size of the state space $|S|$.

\subsection{Algorithm}

The only changes to the \exact algorithm from \cite{bertram} are related to intermediate computation of the value function and subsequent lookup of the value of states which neighbor states that are under consideration during operation of the algorithm.  To differentiate this algorithm from \exact, we name it \algo.

We begin by discussing the algorithm used to compute the value of a state of the intermediate (or final) value function, which requires the (possibly empty) ordered list of peaks that have been processed by previous iterations of the algorithm.  In the event that the list of peaks is empty, a value of 0 is assumed.  Note again that, as with \cite{bertram}, only positive real-valued rewards are considered here.

\begin{algorithm}[H]
\begin{algorithmic}[1]
\Procedure{ValueOnDemand}{previousPeaks, desiredState}
\State $\textit{maxValue} \gets \text{MIN\_FLOAT} $ 

\ForAll{previousPeaks}
	\State $\textit{priValue} \gets pri\_value \times \gamma^{\phi( desiredState, priState)}$
	\State $\textit{secValue} \gets sec\_value \times \gamma^{\phi( desiredState, secState)}$
    \State $\textit{maxValue} \gets \max( maxValue, priValue, secValue ) $ 
\EndFor

\Return $\textit{maxValue}$
\EndProcedure
\end{algorithmic}
\end{algorithm}

The function iterates over all previously selected peaks, keeping track of the maximum value that could be derived from any of the previous peaks, which is the value of the state given the rewards that are represented by the selected peaks.  This is at worst a $O(|R|)$ operation, which grows from $O(1)$ to $O(|R|)$ as the rewards are processed.  Note here that the data structure alluded to here for a peak contains fields for a primary and secondary state. For baseline and delta peaks only the primary is used, for combined peaks both the primary and secondary field are filled in; this is an artifact of implementation details of how the code represents combined peaks.

The remaining changes to the algorithm are simply to replace references to lookup of states in the intermediate value function with calls to this new function, and then removal of the allocation of memory and update of the value function, as shown below.  See the Appendix for full pseudocode.

\begin{algorithm}[H]
\caption{\algo}\label{algorithm}
\begin{algorithmic}[1]
\Procedure{\algo}{$\textit{rewardSources}$}
\State $\textit{processedPeaks} \gets \text{empty list}$
\State $\textit{sortedPeaks} \gets \textit{PrecomputePeaks( rewardSources )}$
\While {$\textit{sortedPeaks} \text{ is not empty}$}
	\State $\textit{deltaPeaks} \gets \textit{ComputeDeltas( processedPeaks )}$
	\State $\textit{sortedPeaks} \gets \textit{PruneInvalidPeaks( processedPeaks)}$
	\State $\textit{maxPeak} \gets \textit{max( [ sortedPeaks, deltaPeaks ] )}$
	\State $\textit{sortedPeaks} \gets \textit{RemoveAffectedPeaks( maxPeak )}$
\EndWhile
\Return $\textit{processedPeaks}$
\EndProcedure
\end{algorithmic}
\end{algorithm}

Line 2 initializes an empty list to track which peaks have been processed by the algorithm.  Line 3 pre-computes baseline peaks and combined peaks based off a list of reward sources and stores them in the form of a sorted list, sorted by value of each peak.  Lines 4-8 continue until we have exhausted the potential peaks and each iteration of the loop whittles away at the list of possible peaks.  Line 5 computes delta peaks for any remaining reward sources by calculating neighboring states values on-demand.  Line 6 removes any peaks that have become invalid due to broken minimum cycles.  Line 7 selects the peak with maximum value.  Line 8 removes any other potential peaks in the list that are affected by selecting the peak with maximum value.  Rather than returning a value function, we instead return the ordered list of peaks that have been processed by the algorithm.

To recover the full value function, we could simply call ValueOnDemand for each state in the state space and then follow the optimal policy normally.  However, we also present a simple algorithm that follows the optimal policy given the final list of peaks produced by the algorithm that does not require computation of the full value function representation.  Starting at the initial state, it computes the value of all neighboring states from the list of peaks.  Once the value of all neighboring states are known, we then have enough information to determine which action is optimal.  We see this as navigating the global value function using only local information about nearby states.

\begin{algorithm}[H]
\caption{FollowLocalPolicy}\label{algorithm}
\begin{algorithmic}[1]
\Procedure{FollowLocalPolicy}{$\textit{processedPeaks, initialState}$}
\State $\textit{currState} \gets \text{initialState}$
\While {$\textit{True}$}
    \State $\textit{neighbor} \gets \textit{FindMaxNeighbor( processedPeaks, currState )}$
    \State $\textit{action} \gets \textit{DetermineAction( currState, neighbor)}$
    \State $\textit{currState} \gets \textit{ExecuteAction( action )}$
\EndWhile
\EndProcedure
\end{algorithmic}
\end{algorithm}

Line 2 initializes the current state, and (due to it being a continuous MDP) lines 3-6 loop forever.  Line 4 finds the maximum neighbor for the current state (which consists of calling ValueOnDemand for each state that can be reached from the current state).  Line 5 determines the (deterministic) action that leads to the neighbor with maximum value (which could be combined with line 4 but is separated here for clarity).  Line 6 represents executing the selected action in the environment and receiving a new state from the environment, which is treated as the current state for the next pass through the loop.

As can be seen from the above, this ends up being a fairly straightforward extension of the prior work once the concepts in \cite{bertram} are understood.  

\newcommand\Rect[3]{
\filldraw[fill=#3!40!white, draw=black] (#1,#2) rectangle (#1+1,#2+1);
}

\newcommand\RArrow[2]{
  \draw[-{Triangle[length=4pt]}, line width=0.5mm]  (#1+.8,#2+.5) -- (#1+1+.2,#2+.5);
}

\newcommand\UArrow[2]{
  \draw[-{Triangle[length=4pt]}, line width=0.5mm]  (#1+.5,#2+.8) -- (#1+.5,#2+1+.2);
}

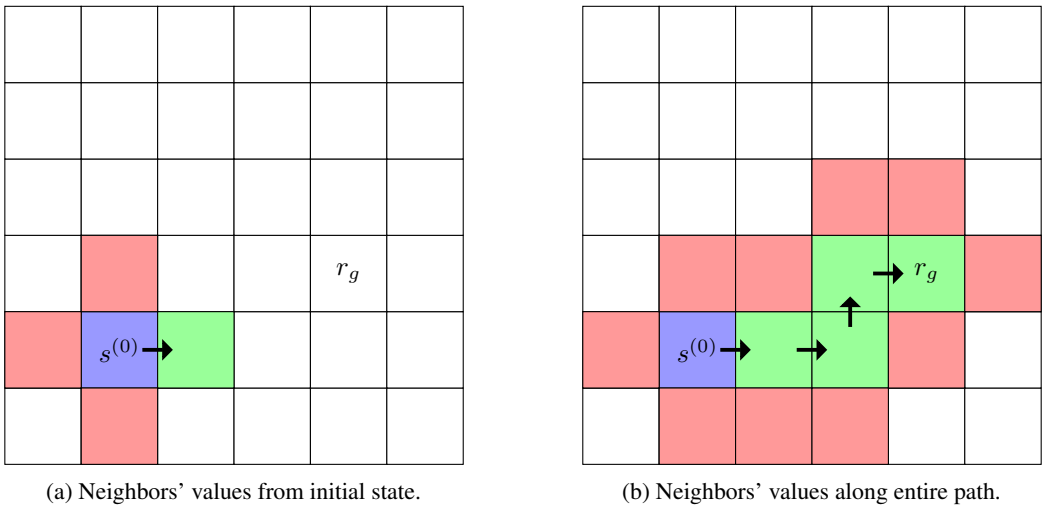
\begin{figure}[H]
\centering
\begin{subfigure}{.45\textwidth}
  \centering
\resizebox{\textwidth}{!}{
\begin{tikzpicture}[node distance = 4cm, auto]
%\draw[step=1cm,gray,very thin] (-1.9,-1.9) grid (5.9,5.9);
 \foreach \x in {0,...,5}
    \foreach \y in {0,...,5} 
        \filldraw[fill=white, draw=black] (\x,\y) rectangle (\x+1,\y+1);

\Rect{1}{1}{blue}
\node (s0) at (1.5,1.5) {$s^{(0)}$};
\node (r) at (4.5,2.5) {$r_g$};

\Rect{2}{1}{green}
\Rect{1}{2}{red}
\Rect{0}{1}{red}
\Rect{1}{0}{red}

%\fill[fill=white] (7,0) rectangle (7+1,0+1);

\RArrow{1}{1}

\end{tikzpicture}
}
  \caption{Neighbors' values from initial state.}
  \label{fig:rewards}
\end{subfigure}\hfill
\begin{subfigure}{.45\textwidth}
  \centering
\resizebox{\textwidth}{!}{
\begin{tikzpicture}[node distance = 4cm, auto]
%\draw[step=1cm,gray,very thin] (-1.9,-1.9) grid (5.9,5.9);
 \foreach \x in {0,...,5}
    \foreach \y in {0,...,5} 
        \filldraw[fill=white, draw=black] (\x,\y) rectangle (\x+1,\y+1);

\Rect{1}{1}{blue}
\node (s0) at (1.5,1.5) {$s^{(0)}$};
\Rect{2}{1}{green}
\Rect{1}{2}{red}
\Rect{0}{1}{red}
\Rect{1}{0}{red}

\Rect{2}{2}{red}
\Rect{2}{0}{red}
\Rect{3}{1}{green}

\Rect{4}{1}{red}
\Rect{3}{0}{red}
\Rect{3}{2}{green}

\Rect{3}{3}{red}
\Rect{4}{2}{green}

\Rect{4}{3}{red}
\Rect{5}{2}{red}

\node (r) at (4.5,2.5) {$r_g$};

%\fill[fill=white] (7,0) rectangle (7+1,0+1);

\RArrow{1}{1}
\RArrow{2}{1}
\UArrow{3}{1}
\RArrow{3}{2}

\end{tikzpicture}

}

  \caption{Neighbors' values along entire path.}
  \label{fig:states}
\end{subfigure}
\caption{Illustration of algorithm calculating neighboring states on-demand as it follows the optimal policy.  The optimal neighbor is shown in green, and the sub-optimal neighbors are shown in red.  The initial state is shown in blue and labeled $s^{(0)}$.  State containing reward labeled $r_g$.  The optimal policy is shown with arrows.  The optimal path is followed by computing the value of only a subset states, where un-colored states are not computed at all.  When the number of states $|S|$ is very large, the number of on-demand computations can be very small compared to the total number of states.}
\label{fig:test}
\end{figure}

\section{Experiments}

\begin{figure}[H]
\centering
\begin{subfigure}{.5\textwidth}
  \centering
  \includegraphics[width=1\linewidth]{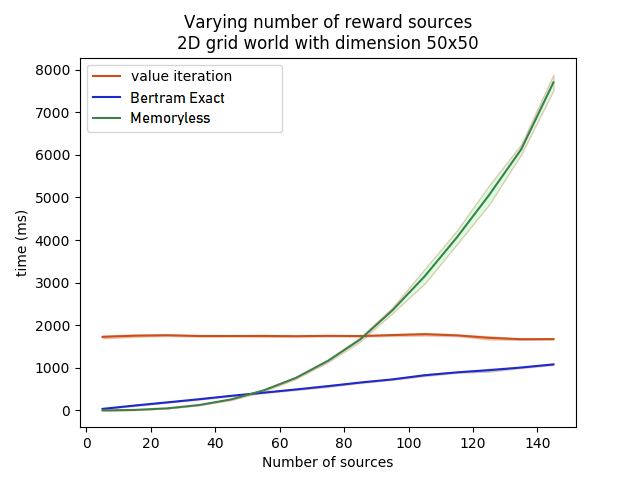}
  \caption{Varying number of reward sources}
  \label{fig:rewards}
\end{subfigure}%
\begin{subfigure}{.5\textwidth}
  \centering
  \includegraphics[width=1\linewidth]{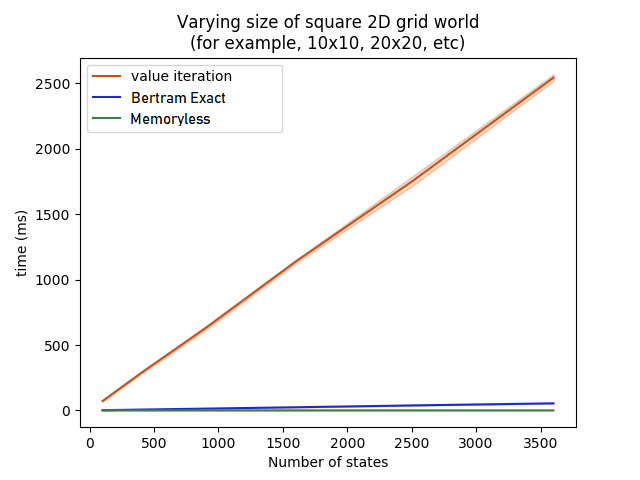}
  \caption{Varying number of states}
  \label{fig:states}
\end{subfigure}
\begin{subfigure}{.5\textwidth}
  \centering
  \includegraphics[width=1\linewidth]{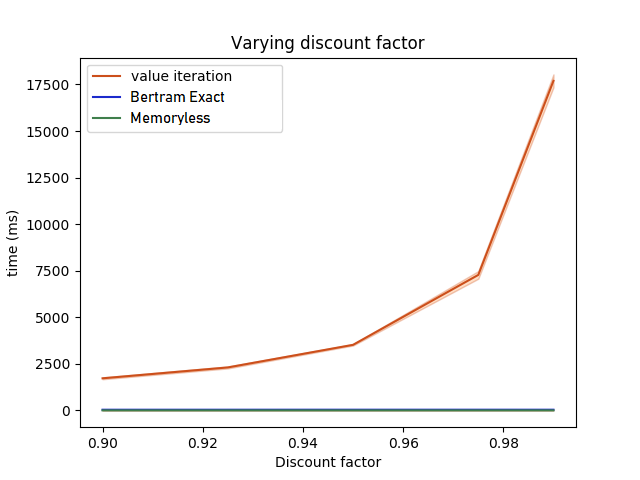}
  \caption{Varying discount factor}
  \label{fig:discount}
\end{subfigure}%
\caption{Experimental results showing performance of the proposed algorithm \algo as compared to value iteration and the \exact algorithm from \cite{bertram}.  (a) For small numbers of rewards, \exact and \algo are comparable in performance.  After a certain point, \algo begins to perform more slowly than both algorithms but recall that \textbf{\algo has no dependency on the size of the state space $|S|$}.  (b) Where \exact had a barely visible dependence on the state space size, \algo is invariant to the number of states.  (c)  Both \exact and \algo remain invariant to the discount factor.}
\label{fig:test}
\end{figure}

Figure \ref{fig:rewards} shows the effects of varying the number of reward sources on the performance of the algorithm.  For this result, a 50x50 grid world was used.  The x-axis shows the number of reward sources used for a test configuration and the y-axis shows the length of time required to solve the MDP.  For each test configuration, 10 randomly generated configurations were created for the number of reward sources specified in the test configuration with reward values ranging from 1 to 10.  For each generated configuration, value iteration, the prior work \exact and this paper's algorithm \algo were run to obtain performance measurements.  As an additional check, the exact solution calculated by this algorithm was compared to the value iteration result to ensure they produced the same result (within a tolerance due to value iteration approximating the exact solution due to the use of a bellman residual as a terminating condition.)  In the plot, the bold line is the average and the colored envelope shows the standard deviation for each test configuration.

The figure shows that as the number of reward sources increases, value iteration remains invariant of the number of reward sources and the prior work grows slowly.  In contrast, we see the tradeoff of increase in time complexity which is traded for not having to hold the value function in memory.  For the algorithm proposed in this paper, for small numbers of reward sources the algorithm clearly continues to outperform value iteration.  As the number of reward sources increases, however, an intersection point will occur and value iteration will begin to perform better.  However, as the size of the state space increases so to does the execution time of value iteration, so the exact point where the intersection occurs will be problem-specific.

Figure \ref{fig:states}  shows the effects of varying the size of the state space on the performance of the algorithm.  For this a fixed number of reward sources (5) were used, and only the size of the state space was varied (by making the grid world larger).  The x axis shows the number of states in the grid world (e.g., $10\times10=100, 50\times50=2500$) and the y axis shows length of time required to solve the MDP.  For each grid world size, 10 randomly generated reward configurations with the fixed number of reward sources were generated.  The results show that value iteration quickly increases in execution time, the prior work grows very slowly, and the algorithm proposed in this paper is invariant of the state space size.

Figure \ref{fig:discount} shows the effects of varying the discount factor on the performance of the algorithm.  For this test, a fixed number of reward sources (5) and state space size (50x50) were used, and only the discount factor was varied.  The x axis shows the discount factor and the y axis shows the length of time required to solve the MDP.  For each discount factor, 10 randomly generated reward configurations with the fixed discount factor were generated.  The results show that value iteration increases apparently exponentially with the discount factor, whereas the prior work and the algorithm proposed in this paper are both invariant to the discount factor.  This follows from the exact calculation of the value based off the distance, where the discount factor is simply a constant that is used in the calculation.

All tests were performed on a high-end "gaming class" Alienware  laptop with a quad-core Intel i7 running at 4.4 GHz with 32GB RAM without using any GPU hardware acceleration (i.e., CPU only).  All code is single threaded, python only and no special optimization libraries other than numpy were used (for example, the python numba library was not used to accelerate numpy calculations.)  Both value iteration and the proposed algorithm use numpy.  The results presented here are meant to most fairly present the performance differences between the algorithms, thus further optimizations should yield improved performance beyond what is presented here.

\section{Conclusion}

In this paper, we have presented a novel approach to solving a certain sub-class of deterministic continuous MDPs exactly that has no dependency on the size of the state space.  This new algorithm's computational speed greatly exceeds that of value iteration for sparse reward sources and, furthermore, is invariant to both the discount factor and the number of states in the state space.  Performance of the algorithm is $O( |R|^3 \times |A|^2  )$, where $|R|$ is the number of reward sources, $|A|$ is the number of actions, and $|S|$ is the number of states.  Memory complexity for the algorithm is $O( |R| \times |A|)$.  We also propose an algorithm to follow the optimal policy using this technique which at each iteration is $O( |R| )$ that leads to an efficient method to both solve the MDP and follow the optimal policy at runtime.  Given the quick time to solve the MDP, it also lends itself to allowing the reward source locations to change arbitrarily between time steps.  Given the lack of dependence on the size of the state space, this algorithm provides a way to solve previously intractable MDPs for which the state-action space was too large to solve exactly.

For deterministic environments with sparse rewards such as certain robotics and unmanned vehicle problems, this new method's performance allows computation to be performed with very minimal memory footprint allowing computations to be performed on very low-performing and low-power embedded hardware.  If the number of rewards is sufficiently small, the algorithm could also perform sufficiently well to allow for real-time constraints to be met in an embedded environment such as a robot or unmanned vehicle.  

To our knowledge, this is the first time that MDPs can be solved exactly without a full representation of the state space held in memory or relying on iterative convergence to the optimal policy or value function.  If this method can be appropriately extended to a larger subset of MDPs (e.g., stochastic MDPs), it could result in broad impacts to the efficiency of solving certain types of MDPs useful in robotics and related spaces.

\bibliographystyle{unsrt}
\bibliography{refs}

\newpage
\section{Appendix 1: Detailed Pseudocode}

\begin{algorithm}[H]
\caption{\algo}\label{algorithm}
\begin{algorithmic}[1]
\Procedure{\algo}{$\textit{rewardSources}$}
\State $\textit{processedPeaks} \gets \text{empty list}$
\State $\textit{sortedPeaks} \gets \textit{PrecomputePeaks( rewardSources )}$
\While {$\textit{sortedPeaks} \text{ is not empty}$}
	\State $\textit{deltaPeaks} \gets \textit{ComputeDeltas( processedPeaks )}$
	\State $\textit{sortedPeaks} \gets \textit{PruneInvalidPeaks( processedPeaks)}$
	\State $\textit{maxPeak} \gets \textit{max( [ sortedPeaks, deltaPeaks ] )}$
	\State $\textit{sortedPeaks} \gets \textit{RemoveAffectedPeaks( maxPeak )}$
\EndWhile
\Return $\textit{processedPeaks}$
\EndProcedure
\end{algorithmic}
\end{algorithm}

Line 2 initializes an empty list to track which peaks have been processed by the algorithm.  Line 3 pre-computes baseline peaks and combined peaks based off a list of reward sources and stores them in the form of a sorted list, sorted by value of each peak.  Lines 4-8 continue until we have exhausted the potential peaks and each iteration of the loop whittles away at the list of possible peaks.  Line 5 computes delta peaks for any remaining reward sources by calculating neighboring states values on-demand.  Line 6 removes any peaks that have become invalid due to broken minimum cycles.  Line 7 selects the peak with maximum value.  Line 8 removes any other potential peaks in the list that are affected by selecting the peak with maximum value.  Rather than returning a value function, we instead return the ordered list of peaks that have been processed by the algorithm.

We next examine the ValueOnDemand function, presenting it out of the calling tree order so that we can characterize its computational complexity to understand its impact on the rest of the code:

\begin{algorithm}[H]
\begin{algorithmic}[1]
\Procedure{ValueOnDemand}{previousPeaks, desiredState}
\State $\textit{maxValue} \gets \text{MIN\_FLOAT} $ 

\ForAll{previousPeaks}
	\State $\textit{priValue} \gets pri\_value \times \gamma^{\phi( desiredState, priState)}$
	\State $\textit{secValue} \gets sec\_value \times \gamma^{\phi( desiredState, secState)}$
    \State $\textit{maxValue} \gets \max( maxValue, priValue, secValue ) $ 
\EndFor

\Return $\textit{maxValue}$
\EndProcedure
\end{algorithmic}
\end{algorithm}

The function iterates over all previously selected peaks, keeping track of the maximum value that could be derived from any of the previous peaks, which is the value of the state given the rewards that are represented by the selected peaks.  This is at worst a $O(|R|)$ operation, which grows from $O(1)$ to $O(|R|)$ as the rewards are processed.  Note here that the data structure alluded to here for a peak contains fields for a primary and secondary state. For baseline and delta peaks only the primary is used, for combined peaks both the primary and secondary field are filled in; this is an artifact of implementation details of how the code represents combined peaks.

\begin{algorithm}[H]
\begin{algorithmic}[1]
\Procedure{PrecomputePeaks}{rewardSources}
\State $\textit{list} \gets \text{empty SortedList} $ 

\ForAll{rewardSources}
	\State $\textit{list.add}$( baseline peak for reward source )
\EndFor

\ForAll{rewardSources}
	\State $nbr \gets \text{find neighboring state with highest reward}$
	\If {nbr is not empty}
	   \State $\textit{list.add}$( cycle peak for reward source )
    \EndIf
\EndFor
\Return $\textit{list}$
\EndProcedure
\end{algorithmic}
\end{algorithm}

Line 2 initializes a sorted list that is sorted by value of the peaks.  In Lines 3-4, a baseline peak is computed for each reward source.  In lines 5-8, if any reward sources are next to each other, their combined peaks are computed.  Note that at this stage, the new ValueOnDemand function is not called; because no peaks have been selected, the value function at this point is assumed to be zeros everywhere.

$\textit{PrecomputePeaks()}$ is a $O(|R| \times |A|)$ algorithm that is done one time at the beginning of the algorithm and yields a list with worst case length of $O(|R| \times |A|)$ entries (but only if the reward sources are all adjacent to each other).  

\begin{algorithm}[H]
\begin{algorithmic}[1]
\Procedure{ComputeDeltas}{processedPeaks}
\State $\textit{list} \gets \text{empty SortedList} $ 

\ForAll{reward sources}
	\State currentValue = $\text{ValueOnDemand( processedPeaks )}$
	\State $\text{compute delta of reward and currentValue }$
	\State $nbr \gets \text{find neighboring state with highest value using ValueOnDemand}$
    
	\State $\textit{list.add}( max( delta peak, neighbor value ) )$
\EndFor
\EndProcedure
\end{algorithmic}
\end{algorithm}

Line 2 initializes a sorted list that is sorted by value of the peaks.  Lines 3-7 compute delta peak for any reward sources that remain.  Lines 4-6 use the new ValueOnDemand function to compute the value of the current and neighboring states.  Line 6-7 properly sort the delta with respect to neighboring states.

$\textit{ComputeDeltas( valueFunction )}$ in \cite{bertram} was a $O(|R|\times |A| )$ algorithm that is done for each pass of the loop, but with the addition of the $O(|R|)$ ValueOnDemand function, the complexity grows to $O(|R|^2 \times |A|)$.

\begin{algorithm}[H]
\begin{algorithmic}[1]
\Procedure{PruneInvalidPeaks}{ processedPeaks}
\ForAll{remaining peaks}
	\State $nbr \gets \text{find neighboring state with highest value using ValueOnDemand}$
	\If {$nbr > peak$}
	    \State $\textit{list.remove( peak )}$
    \EndIf
\EndFor
\EndProcedure
\end{algorithmic}
\end{algorithm}

Lines 2-5 remove any peaks that have become invalid.

$\textit{PruneInvalidPeaks()}$ in \cite{bertram} was a $O(|R| \times |A| )$ algorithm that is done for each pass of the loop.  With the ValueOnDemand function, it now grows to $O(|R|^2 \times |A|)$.

\begin{algorithm}[H]
\begin{algorithmic}[1]
\Procedure{RemoveAffectedPeaks}{list, state}
\ForAll{remaining peaks}
	\If {peak is affected by state}
	    \State $\textit{list.remove( peak )}$
    \EndIf
\EndFor
\EndProcedure
\end{algorithmic}
\end{algorithm}

Lines 2-4 remove any peaks that have been eliminated by the choice of the peak with maximum value.

$\textit{RemoveAffectedPeaks}$ operates over the $O(|R| \times |A|)$ $\textit{sortedPeaks}$ list, but this also shrinks by $O(|A|)$ entries each pass.

\subsubsection{Time Complexity}

The main loop of the \algo function is a $O(|R| \times |A|)$ function, but the ComputeDelta and PruneInvalidPeaks functions are both $O( |R|^2 \times |A| )$ due to their usage of the ValueOnDemand function, bringing the overall algorithm complexity to $O(|R|^3 \times |A|^2)$.  Note here there is no dependence upon the size of the state space $|S|$.

For environments where the connected distance is not easily determined (arbitrary transition graph), then the complexity to determine the distance between states must be taken into consideration.  However, it is assumed that this can be precomputed offline because $T$ is assumed to be stationary. 

For environments like the 2D grid world where the structure of the space is known, determining the connected distance between states is a $O(1)$ calculation, which can be represented as a simple function call to determine the neighbors of each state on-demand.

\subsubsection{Memory Complexity}

Memory complexity for the algorithm is $O( |R| \times |A|)$

\end{document}